\documentclass[11pt]{article}
\usepackage{coling2020}
\usepackage{times}
\usepackage{url}
\usepackage{latexsym}

\usepackage{amsmath}
\usepackage{multirow}
\usepackage{danudefs}
\usepackage{algorithmic}
\usepackage{algorithm}
\usepackage{graphicx}
\graphicspath{{img/}}
\usepackage{epstopdf}
\usepackage{multirow}
\usepackage{booktabs}
\usepackage{subcaption}
\usepackage[normalem]{ulem}
\usepackage{enumitem}
\usepackage{float}
\usepackage{caption}
\usepackage{subcaption}
\usepackage{framed}
\usepackage{hyperref}

\newcommand{\require}{\textbf{Input: }}
\newlength\mylen

\newcommand{\ensure}{\textbf{Output: }}

\usepackage[disable]{todonotes} 

\makeatletter
\if@todonotes@disabled

\else

\fi
\makeatother

 \newcommand\DB[2][]{\todo[inline, size=\small, caption={2do}, #1]{
\begin{minipage}{\textwidth-4pt}DB: #2\end{minipage}}}

\renewcommand{\ss}{\scriptscriptstyle}

\title{Multi-Source Attention for Unsupervised Domain Adaptation}

 \author{Xia Cui \\
   University of Liverpool, \\
   United Kingdom.\\
   {\tt Xia.Cui@liverpool.ac.uk} \\\And
   Danushka Bollegala \\
   University of Liverpool \\
   United Kingdom \\
   {\tt danushka@liverpool.ac.uk} \\}

\date{}

\begin{document}
\maketitle
\begin{abstract}
	Domain adaptation considers the problem of generalising a model learnt using data from a particular source domain to a different target domain. 
Often it is difficult to find a suitable single source to adapt from, and one must consider multiple sources.
Using an unrelated source can result in sub-optimal performance, known as the \emph{negative transfer}.
However, it is challenging to select the appropriate source(s) for classifying a given target instance in multi-source unsupervised domain adaptation (UDA).
We model source-selection as an attention-learning problem, where we learn attention over sources for a given target instance.
For this purpose, we first independently learn source-specific classification models, and a relatedness map between sources and target domains using pseudo-labelled target domain instances. 
Next, we learn attention-weights over the sources for aggregating the predictions of the source-specific models.
Experimental results on cross-domain sentiment classification benchmarks show that the proposed method outperforms prior proposals in multi-source UDA.\footnote{Source code available at \url{https://github.com/summer1278/multi-source-attention}}
\end{abstract}

\section{Introduction}
\label{sec:intro}

Many machine learning processes have different training and testing distributions~\cite{Zhang:AAAI:2015}, thus leading to the problem of Domain Adaptation (DA). 
Most DA methods consider adapting to a target domain from a single source domain~\cite{Blitzer:EMNLP:2006,Blitzer:ACL:2007,Ganin:JMLR:2016}. 
The goal of DA is to transfer salient information from the source domain to obtain a model suitable for a given target domain~\cite{Cheng:IEICE:2014}.
In practice, however, training data can come from multiple sources.
 For example, in sentiment classification, each product category is considered as a \emph{domain}~\cite{Blitzer:EMNLP:2006}, resulting in a multi-domain adaptation setting.

Unsupervised DA~(UDA) is a special case of DA where labelled instances are not available for the target domain.
Existing approaches for UDA can be categorised into pivot-based and instance-based methods.
Pivots refer to the features common to both source and target domain~\cite{Blitzer:EMNLP:2006}.
Pivot-based single-source domain adaptation methods, such as Structural Correspondence Learning ~(SCL) \cite{Blitzer:EMNLP:2006,Blitzer:ACL:2007}
and Spectral Feature Alignment~(SFA) \cite{Pan:WWW:2010}, first select a set of pivots and then project the source and target domain documents into a shared space. 
Next, a prediction model is learnt in this shared space.
However, these methods fail in multi-source settings because it is challenging to find pivots across all sources such that a shared projection can be learnt.
Similarly, instance-based methods, such as Stacked Denoising Autoencoders~(SDA) \cite{Glorot:ICML:2011} and marginalized Stacked Denoising Autoencoders~(mSDA) \cite{Chen:ICML:2012} minimise the loss between the original inputs and their reconstructions. 
Not all of the source domains are appropriate for learning transferable projections for a particular target domain. 
Adapting from an unrelated source can result in poor performance on the given target, which is known as \textit{negative transfer}~\cite{Rosenstein:NIPS:2005,Pan:TKDE:2010,Guo:EMNLP:2018}.

Prior proposals for multi-source UDA can be broadly classified into methods that:
(a) first select a source domain and then select instances from that source domain to adapt to a given target domain test instance~\cite{Ganin:JMLR:2016,Kim:ACL:2017,Zhao:NIPS:2018,Guo:EMNLP:2018};
(b) pool all source domain instances together and from this pool select instances to adapt to a given target domain test instance~\cite{Chattopadhyay:TKDD:2012};
(c) pick a source domain and use all instances in that source (source domain selection)~\cite{Schultz:arXiv:2018}; and
(d) pick all source domains and use all instances (utilising all instances)~\cite{Aue:RANLP:2005,Bollegala:ACL:2011,Wu:ACL:2016}.

In contrast, in this paper, we propose a multi-source UDA method an make the following contributions:
\begin{itemize}[noitemsep,topsep=0pt]
    \item We propose a self-training-based pseudo-labelling method for learning an attention model for multi-source UDA. The proposed method learns \emph{domain-attention} weights for the source domains per test instance. Based on the learnt attention scores, we are able to find appropriate sources to adapt to a given target domain.
    \item Unlike adversarial neural networks based approaches~\cite{Ganin:JMLR:2016,Guo:EMNLP:2018}, our proposed method does not require rule-based labelling of instances for training.
    \item We evaluate the performance of the proposed method against pivot- and instance-based approaches. The proposed method performs competitively against previously proposed multi-source UDA methods and is able to provide evidence for its predictions.
\end{itemize}



\section{Related Work}

In Section~\ref{sec:intro} we mentioned prior proposals for single-source DA and this section discusses multi-source DA, which is the main focus of this paper.
\newcite{Bollegala:ACL:2011} created a sentiment sensitive thesaurus~(SST) using the data from the union of multiple source domains to train a cross-domain sentiment classifier. The SST is used to expand feature spaces during train and test times. The performance of their method depends heavily on the selection of pivots~\cite{Xia:ECML:2017,Li:IJCAI:2017}. 
\newcite{Wu:ACL:2016} proposed a sentiment DA method from multiple sources (SDAMS) by introducing two components: a sentiment graph and a domain similarity measure. The sentiment graph is extracted from unlabelled data. 
Similar to SST, SDAMS utilises data from multiple sources to maximise the available labelled data.
\newcite{Guo:AAAI:2020} proposed a mixture of distance measures including and used a multi-arm bandit to dynamically select a single source during training. 
However, in our proposed method all domains are selected and contributing differently as specified by their domain-attention weights for each train and test instance. 
Moreover, we use only one distance measure and is conceptually simple to implement.

Recently, Adversarial NNs have become popular in DA~\cite{Ganin:JMLR:2016,Zhao:NIPS:2018,Guo:EMNLP:2018}. 
Adversarial training is used to reduce the discrepancy between source and target domains~\cite{Ding:IEEE:2019}.
\newcite{Ganin:JMLR:2016} proposed Domain-Adversarial Neural Networks (DANN) that use a gradient reversal layer to learn domain independent features for a given task. 
\newcite{Zhao:NIPS:2018} proposed Multiple Source Domain Adaptation with Adversarial Learning (MDAN), a generalisation of DANN that aims to learn domain independent features while being relevant to the target task. 
\newcite{Li:IJCAI:2017} proposed End-to-End Adversarial Memory Network (AMN), inspired by memory networks~\cite{sukhbaatar:NIPS:2015}, and automatically capture pivots using an attention mechanism. 
\newcite{Guo:EMNLP:2018} proposed an UDA method using a mixture of experts for each domain. They model the domain relations using a \textit{point-to-set} distance metric to the encoded training matrix for source domains. Next, they perform joint training over all domain-pairs to update the parameters in the model by \textit{meta-training}. However, they ignore the available unlabelled instances for the source domain. Adversarial training methods have shown to be sensitive to the hyper parameter values and require problem-specific techniques~\cite{Mukherjee:EMNLP:2018}. 
\newcite{Kim:ACL:2017} models domain relations using \textit{example-to-domain} based on an attention mechanism. However, the attention weights are learnt using source domain training data in a supervised manner.

Following a self-training approach, \newcite{Chattopadhyay:TKDD:2012} proposed a two-stage weighting framework for multi-source DA that first computes the weights for features from different source domains using Maximum Mean Discrepancy (MMD) \cite{Borgwardt:BioInfo:2006}.
Next, they generate pseudo labels for the target unlabelled instances using a classifier learnt from the multiple source domains. Finally, a classifier is trained on the pseudo-labelled instances for the target domain. 
Their method requires labelled data for the target domain, which is a \textit{supervised} DA setting, different from the UDA setting we consider in this paper.
Our proposed method uses self-training to assign pseudo-labels for the unlabelled target instances, and learn an embedding for each domain using an attention mechanism.
\section{Multi-Source Domain Attention}
Let us assume that are given $N$ source domains, $S_{1}, S_{2}, \ldots, S_{N}$, and required to adapt to a target domain $T$.
Moreover, let us denote the labelled instances in $S_{i}$ by $\cS_{i}^{\ss L}$ and unlabelled instances by $\cS_{i}^{\ss U}$.
For $T$ we have only unlabelled instances $\cT^{U}$ in the UDA setting.
Our goal is to learn a classifier to predict labels for the target domain instances using $\cS^{\ss L} = \cup_{i=1}^{N} \cS_{i}^{\ss L}$, $\cS^{\ss U} = \cup_{i=1}^{N} \cS^{\ss U}_{i}$ and $\cT^{\ss U}$.
We denote labelled and unlabelled instances in $S_{i}$ by respectively $x_{i}^{\ss L}$ and $x_{i}^{\ss U}$, whereas instances in $T$ are denoted by $x_{\ss T}$.
To simplify the notation, we drop the superscripts $L$ and $U$ when it is clear from the context whether the instance is respectively labelled or not.

The steps of our proposed method can be summarised as follows: 
(a) use labelled and unlabelled instances from each of the source domains to learn classifiers that can predict the label for a given instance.  
Next, develop a majority voter and use it to predict the \emph{pseudo-labels} for the target domain unlabelled instances $\cT^{\ss U}$ (Section~\ref{sec:pseudo-label});
(b) compute a \textit{relatedness map} between the target domain's pseudo-labelled instances, $\cT^{\ss L*}$, and source domains' labelled instances $\cS^{\ss L}$ (Section~\ref{sec:relate-map});
(c) compute \emph{domain-attention} weights for each source domain (Section~\ref{sec:instance-att});
(d) jointly learn a model based on the relatedness map and the domain-attention weights for predicting labels for the target domain's test instances (Section~\ref{sec:training}).

\subsection{Pseudo-Label Generation}
\label{sec:pseudo-label}
In UDA, we have only unlabelled data for the target domain. 
Therefore, we first introduce pseudo-labels for the target domain instances $\cT^{\ss U}$ by self-training~\cite{Abney:SLC:2007} following Algorithm~\ref{alg:MS-self-train}.
Specifically, we first train a predictor $f_{i}$ for the $i$-th source domain using only its labelled instances $\cS_{i}^{\ss L}$ using a base learner $\Gamma$ (Line~1-2).
Any classification algorithm that can learn a predictor $f_{i}$ that can compute the probability, $f_{i}(x,y)$, of a given instance $x$ belonging to the class $y$ can be used as $\Gamma$.
In our experiments, we use logistic regression for its simplicity and popularity in prior UDA work~\cite{Bollegala:ACL:2011,Bollegala:IEEE:2013}.
Next, for each unlabelled instance in the selected source domain, we compute the probability of it belonging to each class and find the most probable class label.
If the probability of the most likely class is greater than the given confidence threshold $\tau \in [0,1]$, we will append that instance to the current labelled training set.
This enables us to increase the labelled instances for the source domains, which is important for learning accurate classifiers when the amount of labelled instances available is small.
After processing all unlabelled instances in domain $S_{i}$ we train the final classifier $f_{i}$ for that domain using all initial and pseudo-labelled instances.
We predict a pseudo-label for a target domain instance as the majority vote, $f^*$, over the predictions made by the individual classifiers $f_{i}$. 

{\small
\begin{algorithm}[tbh] 
	\caption{Multi-Source Self-Training} 
	\label{alg:MS-self-train} 
	\require source domains' labelled instances $\cS_{1}^{\ss L}, \ldots, \cS_{N}^{\ss L}$, source domains' unlabelled instances $ \cS_{1}^{\ss U}, \ldots, \cS_{N}^{\ss U}$ and target domain's unlabelled instances $\cT^{\ss U}$, target classes $\cY$, base learner $\Gamma$ and the classification confidence threshold $\tau$.
	
	\ensure multi-source self-training classifier $f^*$
	
	\begin{algorithmic}[1] 
	    \FOR{$i =1~\TO~N$}
	    \STATE{$\cL_i \leftarrow \cS_i^{\ss L}$}
	    	\STATE{$ f_{i} \leftarrow \Gamma(\cL_{i})$}
		\FOR{$ x \in \cS_{i}^{\ss U}$}
		\STATE{$\hat{y} = \arg\max_{y \in \cY} f_{i}(x, y)$}
		\IF{$f_{i}(x, \hat{y}) > \tau$}
			\STATE{$\cL_i \leftarrow \cL_i \cup \{(x, \hat{y})\}$}
		\ENDIF
		\ENDFOR
		 \STATE${f_{i} \leftarrow \Gamma(\cL_{i})}$
		\ENDFOR
		\RETURN majority voter $f^*$ over $f_1, \ldots, f_{N}$.
	\end{algorithmic}
\end{algorithm}
}

Selecting the highest confident pseudo-labelled instances for the purpose of training a classifier for the target domain has been a popular as done in prior work~\cite{Zhou:TKDE:2005,Abney:SLC:2007,Sogaard:ACL:2010,Ruder:ACL:2018} does not guarantee that those instances will be the most suitable ones for adapting to the target domain, which was not considered during the self-training stage.
For example, some target instances might not be good prototypical examples of the target domain and we would not want to use the pseudo-labels induced for those instances when training a classifier for the target domain.
To identify instances in the target domain that are better prototypes, we first encode each target instance by a vector and select the instances that are closest to the centroid, $\vec{c}_{\scriptscriptstyle T}$, of the target domain instances given by \eqref{eq:centroid}.
\begin{align}
	\label{eq:centroid}
    \vec{c}_{\scriptscriptstyle T} = \dfrac{1}{|\cT^{\ss U}|} \sum_{x \in \cT^{\ss U}} \vec{x}
\end{align}
In the case of text documents $x$, their embeddings, $\vec{x}$,  can be computed using numerous approaches such as using bi-directional LSTMs~\cite{melamud-etal-2016-context2vec} or transformers~\cite{reimers-gurevych-2019-sentence}.
In our experiments, we use the Smoothed Inversed Frequency~(SIF) proposed by \newcite{Arora:ICLR:2017}, which computes document embeddings as the weighted-average of the pre-trained word embeddings for the words in a document.
Despite being unsupervised, SIF has shown strong performance in numerous semantic textual similarity benchmarks~\cite{agirre-etal-2015-semeval}.
Using the centroid computed in \eqref{eq:centroid}, similarity for target instance to the centroid is computed using the cosine similarity given in \eqref{eq:tgt-sim}.
\begin{align}
    \textrm{sim}(\vec{x}, \vec{c}_{\ss T}) = \dfrac{\vec{x}\T \vec{c}_{\ss T}}{\norm{\vec{x}} \norm{\vec{c}_{\ss T}}}
    \label{eq:tgt-sim}
\end{align}
Other distance measures such as the Euclidean distance can also be used. We use cosine similarity here for its simplicity. 
We predict the labels for the target domain unlabelled instances, $\cT^{\ss U}$, using $f^*$, and select the instances with the top-$k$ highest similarities to the target domain according to \eqref{eq:tgt-sim} as the target domain's pseudo-labelled instances $\cT^{\ss L*}$.

\subsection{Relatedness Map Learning}
\label{sec:relate-map}

Not all of the source domain instances are relevant to a given target domain instance and the performance of a classifier under domain shift can be upper bounded by the $\cH$-divergence between a source and a target domain~\cite{Kifer:2004,David:NIPS:2006,Blitzer:ML:2009}.
To model the relatedness between a target domain instance and each instance from the $N$ source domains, we use the pseudo-labelled target domain instances $\cT^{L*}$ and source domains' labelled instances $\cS_{i}^{L}$ to learn a \emph{relatedness map}, $\psi_{i}$, between a target domain instance $\vec{x}_{\ss T} (\in  \cT^{L*}$) and a source domain labelled instance $\vec{x}_{i}^{\ss L}$ ($\in \cS_{i}^{L}$) as given by \eqref{eq:related-map}.
\begin{align}
	\psi_{i}(\vec{x}_{\ss T}, \vec{x}_{i}^{\ss L}) = \dfrac{\exp(\vec{x}_{\ss T}\T \vec{x}_{i}^{\ss L})}{\sum_{\vec{x}' \in \cS_{i}^{\ss L}}{\exp(\vec{x}_{\ss T}\T\vec{x}')}}
	\label{eq:related-map}
\end{align}
With the help of the relatedness map, $\psi_{i}$, we can determine how well each instance in a source domain contributes to the prediction of the label of a target domain's instance.

\subsection{Instance-based Domain-Attention}
\label{sec:instance-att}


To avoid negative transfer, we dynamically select the source domain(s) to use when predicting the label for a given target domain instance.
Specifically, we learn \emph{domain-attention}, $\theta(\vec{x}_{\ss T},\cS_i)$, for each source domain, $S_{i}$, conditioned on $x_{\ss T}$ as given by~\eqref{eq:domain-att}.
\begin{align}
	\theta(\vec{x}_{\ss T},\cS_i) = \dfrac{\exp(\vec{x}_{\ss T}\T \vec{\phi}_i)}{\sum_{j=1}^N{\exp(\vec{x}_{\ss T}\T \vec{\phi}_j)}}
	\label{eq:domain-att}
\end{align}
$\vec{\phi}_i$ can be considered as a \textit{domain embedding} for $S_{i}$ and has the same dimensionality as the instance embeddings.
During training, to prevent activation outputs from exploding or vanishing, we initialise $\vec{\phi}_i$ using Xavier initialisation~\cite{Glorot:ICAI:2010} and normalise such that $\forall \vec{x}_{\ss T}, \sum_{i=1}^N{\theta(\vec{x}_{\ss T},\cS_i)}=1$. 

\subsection{Training}
\label{sec:training}
We combine the relatedness map (Section~\ref{sec:relate-map}) and domain-attention (Section~\ref{sec:instance-att}) and predict the label, $\hat{y}(x_{\ss T})$, of a target domain instance $x_{\ss T}$ using \eqref{eq:prediction}.
\begin{align}
	\hat{y}(x_{\ss T}) = \sigma \left(\sum^{N}_{i=1} \sum_{\vec{x}_{i}^{\ss L} \in \cS_{i}^{\ss L}} y(x_{i}^{\ss L}) \, \psi_{i}(\vec{x}_{\ss T}, \vec{x}_{i}^{\ss L}) \, \theta(\vec{x}_{\ss T},\cS_i) \right)
	\label{eq:prediction}
\end{align}
Here, $\sigma(z) = 1/(1+\exp(-z))$ is the logistic sigmoid function and $y(x_{i}^{\ss L})$ is the label of the source domain labelled instance $x_{i}^{\ss L}$.

First, we use the target instances, $x \in \cT^{L*}$, with inferred labels $y^{*}(x)$ (computed using $f^{*}$ produced by Algorithm~\ref{alg:MS-self-train}) as the training instances and predict their labels, $\hat{y}(x)$, by \eqref{eq:prediction}.
The cross entropy error, $E\left(\hat{y}(x), y^{*}(x)\right)$ for this prediction is given by~\eqref{eq:pred-loss}:
\begin{align}
	E\left(\hat{y}(x), y^{*}(x)\right) = - \lambda(x) (1 - y^*(x)) \log(1-\hat{y}(x))- \lambda(x) y^{*}(x) \log (\hat{y}(x))
	\label{eq:pred-loss}
\end{align}
Here, $\lambda(x)$ a rescaling factor computed using the normalised similarity score as in~\eqref{eq:norm-tgt-sim}:
\begin{align}
    \lambda(x) = \dfrac{\textrm{sim}(\vec{x}, \vec{c}_{\ss T})}{\sum_{\vec{x}' \in  \cT^{L*}}{\textrm{sim}(\vec{x}', \vec{c}_{\ss T})}}
    \label{eq:norm-tgt-sim}
\end{align}
We minimise the cross-entropy error given by \eqref{eq:pred-loss} using ADAM~\cite{Kingma:ICLR:2015} for the purpose of learning the domain-embeddings, $\vec{\phi}_{i}$.
The initial learning rate in ADAM was set to $10^{-3}$ using a subset of $\cT^{L*}$ held-out as a validation dataset.

\section{Experiments}
To evaluate the proposed method, we use the multi-domain Amazon product review dataset compiled by~\newcite{Blitzer:ACL:2007}. 
This dataset contains product reviews from four domains: Books (\textbf{B}), DVD (\textbf{D}), Electronics (\textbf{E}) and Kitchen Appliances (\textbf{K}). 
Following~\newcite{Guo:EMNLP:2018}, we conduct experiments under two different splits of this dataset as originally proposed by~\newcite{Blitzer:ACL:2007}~(\textbf{Blitzer2007}) and by~\newcite{Chen:ICML:2012}~(\textbf{Chen2012}). 
Table~\ref{tbl:num-reviews-amazon} shows the number of instances in each dataset.
By using these two versions of the Amazon review dataset, we can directly compare the proposed method against relevant prior work.
Next, we describe how the proposed method was trained on each dataset.

For \textbf{Blitzer2007}, we use the official train and test splits where each domain contains $1600$ labelled training instances ($800$ positive and $800$ negative), and $400$ target test instances ($200$ positive and $200$ negative).
    In addition, each domain also contains 6K-35K unlabelled instances.
    We use 300 dimensional pre-trained GloVe embeddings~\cite{Pennington:EMNLP:2014} following prior work~\cite{Bollegala:ACL:2011,Wu:ACL:2016} with SIF \cite{Arora:ICLR:2017} to create document embeddings for the reviews.
    
    
 In \textbf{Chen2012}, each domain contains $2000$ labelled training instances ($1000$ positive and $1000$ negative), and $2000$ target test instances ($1000$ positive and $1000$ negative). 
    The remainder of the instances are used as unlabelled instances (ca. 4K-6K for each domain).
    We use the publicly available\footnote{\url{https://github.com/KeiraZhao/MDAN/}} $5000$ dimensional tf-idf vectors produced by~\newcite{Zhao:NIPS:2018}.
    We use a multilayer perceptron (MLP) with an input layer of $5000$ dimensions and $3$ hidden layers with $500$ dimensions. 
    We use final output layer with $500$ dimensions as the representation of an instance.

For each setting, we follow the standard input representation methods as used in prior work. 
It also shows the flexibility of the proposed method to use different (embedding vs. BoW) text representation methods.
We conduct experiments for cross-domain sentiment classification with multiple sources by selecting one domain as the target and the remaining three as sources. 
The statistics for the two settings are shown in Table~\ref{tbl:num-reviews-amazon}.

\begin{table}[t]
\small
	\centering
	\resizebox{0.8\textwidth}{!}{%
\begin{tabular}{ll|lll|lll}
\toprule
Target    & Source   & Train                  & Test          & Unlabel         & Train                & Test       & Unlabel       \\ 
\multicolumn{2}{c|}{} & \multicolumn{3}{l|}{Blitzer2007~\cite{Blitzer:EMNLP:2006}} & \multicolumn{3}{l}{Chen2012~\cite{Chen:ICML:2012}} \\ \midrule
B         & D,E,K    & $1600\times3$          & 400           & 6000            & $2000\times3$        & 2000       & 4465          \\
D         & B,E,K    & $1600\times3$          & 400           & 34741           & $2000\times3$        & 2000       & 5586          \\
E         & B,D,K    & $1600\times3$          & 400           & 13153           & $2000\times3$        & 2000       & 5681          \\
K         & B,D,E    & $1600\times3$          & 400           & 16785           & $2000\times3$        & 2000       & 5945   
\\ \bottomrule
\end{tabular}
	}
	\caption{Number of train, test and unlabelled instances for the two Amazon product review datasets.}
	\label{tbl:num-reviews-amazon}
\end{table}

\subsection{Comparisons against Prior Work}
We evaluate the proposed method in two settings. 
In Table~\ref{tbl:Blitzer2007-acc}, we compare our method against the following methods on \textbf{Blitzer2007} dataset:
\begin{description}[leftmargin=*]
\itemsep=-3pt
\item[uni-MS:] is the baseline model, trained on the union of all source domains and tested directly on a target domain without any DA. uni-MS has been identified as a strong baseline for multi-source DA~\cite{Aue:RANLP:2005,Zhao:NIPS:2018,Guo:EMNLP:2018}. 
\item[SCL:] Structural Correspondence Learning~\cite{Blitzer:EMNLP:2006,Blitzer:ACL:2007} is a single-source DA method, trained on the union of all source domains and tested on the target domain. We report the published results from~\newcite{Wu:ACL:2016}.
\item[SFA:] Spectral Feature Alignment~\cite{Pan:WWW:2010} is a single-source DA method, trained on the union of all source domains, and tested on the target domain.  We report the published results from~\newcite{Wu:ACL:2016}.
\item[SST:] Sensitive Sentiment Thesaurus~\cite{Bollegala:ACL:2011,Bollegala:IEEE:2013} is the SoTA multi-source DA method on \textbf{Blitzer2007}.  We report the published results from \newcite{Bollegala:ACL:2011}.
\item[SDAMS:] Sentiment Domain Adaptation with Multiple Sources proposed by \newcite{Wu:ACL:2016}. We report the results from the original paper.
 \item [AMN:] End-to-End Adversarial Memory Network~\cite{Li:IJCAI:2017} is a single-source DA method, trained on the union of all source domains, and tested on the target domain. We report the published results from \newcite{Ding:IEEE:2019}.
 \end{description}

\begin{table}[ht]
\centering
\resizebox{0.6\textwidth}{!}{
\begin{tabular}{@{}llllllll@{}}
\toprule
T & uni-MS & SCL   & SFA   & SST   & SDAMS & AMN   & Proposed \\ \midrule
B      & 80.00     & 74.57 & 75.98 & 76.32 & 78.29 & 79.75 & \textbf{83.50}     \\
D      & 76.00     & 76.30  & 78.48 & 78.77 & 79.13 & 79.83 & \textbf{80.50}     \\
E      & 74.75  & 78.93 & 78.08 & 83.63* & \textbf{84.18**} & 80.92* & 80.00*       \\
K      & 85.25  & 82.07 & 82.10  & 85.18 & \textbf{86.29} & 85.00    & 86.00     \\ \bottomrule
\end{tabular}
}
\caption{Classification accuracies (\%) for the proposed method and prior work on \textbf{Blitzer2007}. Statistically significant improvements over \textbf{uni-MS} according to the Binomial exact test are shown by ``*'' and ``**'' respectively at $p=0.01$ and $p=0.001$ levels.}
\label{tbl:Blitzer2007-acc}
\end{table}

In Table~\ref{tbl:chen2012-acc}, we compare our proposed method against the following methods on \textbf{Chen2012}.
\begin{description}[leftmargin=*]
\itemsep=-3pt
 \item[mSDA:] Marginalized Stacked Denoising Autoencoders proposed by~\newcite{Chen:ICML:2012}. We report the published results from~\newcite{Guo:EMNLP:2018}.
 \item[DANN:] Domain-Adversarial Neural Networks proposed by~\newcite{Ganin:JMLR:2016}. We report the published results from \newcite{Zhao:NIPS:2018}.
\item[MDAN:] Multiple Source Domain Adaptation with Adversarial Learning proposed by~\newcite{Zhao:NIPS:2018}.  We report the published results from the original paper.
 \item[MoE:] Mixture of Experts proposed by~\newcite{Guo:EMNLP:2018}. We report the published results from the original paper.
 \end{description}

\begin{table}[htb]
\centering
\resizebox{0.55\textwidth}{!}{
\begin{tabular}{@{}lllllll@{}}
\toprule
T & uni-MS & mSDA  & DANN  & MDAN  & MoE   & Proposed \\ \midrule
B      & 79.46  & 76.98 & 76.50  & 78.63 & 79.42 & \textbf{79.68}    \\
D      & 82.32  & 78.61 & 77.32 & 80.65 & \textbf{83.35} & 82.96    \\
E      & 84.93  & 81.98 & 83.81 & 85.34 & \textbf{86.62} & 85.30     \\
K      & 86.71  & 84.26 & 84.33 & 86.26 & \textbf{87.96} & 87.48    \\ \bottomrule
\end{tabular}
}
\caption{Classification accuracies (\%) for the proposed method and prior work on \textbf{Chen2012}.}
\label{tbl:chen2012-acc}
\end{table}


From Table~\ref{tbl:Blitzer2007-acc} and Table~\ref{tbl:chen2012-acc}, we observe that the proposed method obtains the best classification accuracy on Books domain (\textbf{B}) in both settings, which is the domain with the smallest number of unlabelled instances.
\DB{Can we tell anything else about Tables 3 and 4 in addition to B being the best?}

\subsection{Effect of Self-Training}
As described in Section~\ref{sec:pseudo-label}, our proposed method uses self-training to generate pseudo-labels for the target domain unlabelled instances. 
In Table~\ref{tbl:chen2012-self-train}, we compare self-training against alternative pseudo-labelling methods on \textbf{Chen2012}:
Self-Training~(\textbf{Self}) \cite{Abney:SLC:2007,Chattopadhyay:TKDD:2012}, Union Self-Training~(\textbf{uni-Self}) \cite{Aue:RANLP:2005}, Tri-Training~(\textbf{Tri}) \cite{Zhou:TKDE:2005} and Tri-Training with Disagreement~(\textbf{Tri-D}) \cite{Sogaard:ACL:2010}.
In Table~\ref{tbl:chen2012-self-train}, we observe that all semi-supervised learning methods improve only slightly over uni-MS (no adapt baseline).
Therefore, pseudo-labelling step alone is insufficient for DA. 
Moreover, we observe that all semi-supervised methods perform comparably.

\begin{table}[htb]
\centering
\resizebox{0.45\textwidth}{!}{
\begin{tabular}{llllll}
\toprule
T & uni-MS & Self  & uni-Self & Tri   & Tri-D \\ \midrule
B & 79.46  & 79.60  & 79.46      & \textbf{79.61} & 79.51 \\
D & 82.32  & \textbf{82.49} & 82.35      & 82.35 & 82.35 \\
E & 84.93  & 84.97 & 84.93      & \textbf{84.99} & 84.93 \\
K & 87.17  & 87.18 & 87.17      & 87.15 & \textbf{87.23} \\ \bottomrule
\end{tabular}
}
\caption{Classification accuracies (\%) for semi-supervised methods on \textbf{Chen2012}.}
\label{tbl:chen2012-self-train}
\end{table}

\begin{figure*}[htb]
\centering
    \begin{subfigure}[b]{0.4\textwidth}
	    \includegraphics[width=\textwidth]{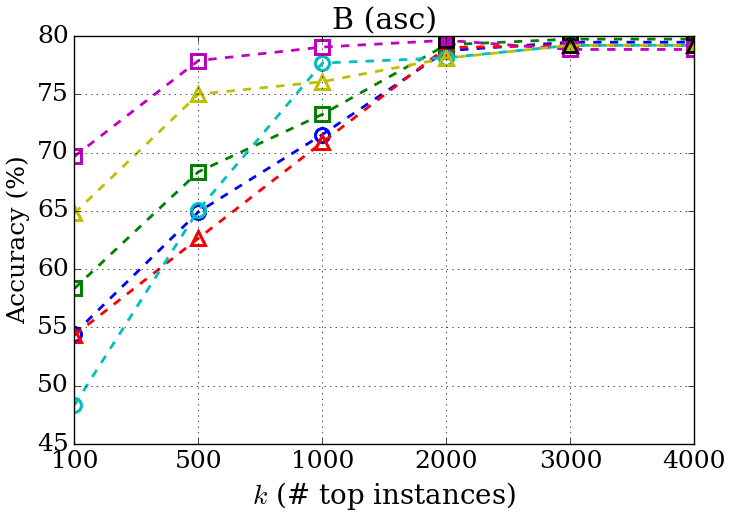}
	    \caption{prob sorted in ascending order}
	    \label{fig:selection-methods-asc}
	\end{subfigure}
	\begin{subfigure}[b]{0.5\textwidth}
	\includegraphics[width=\textwidth]{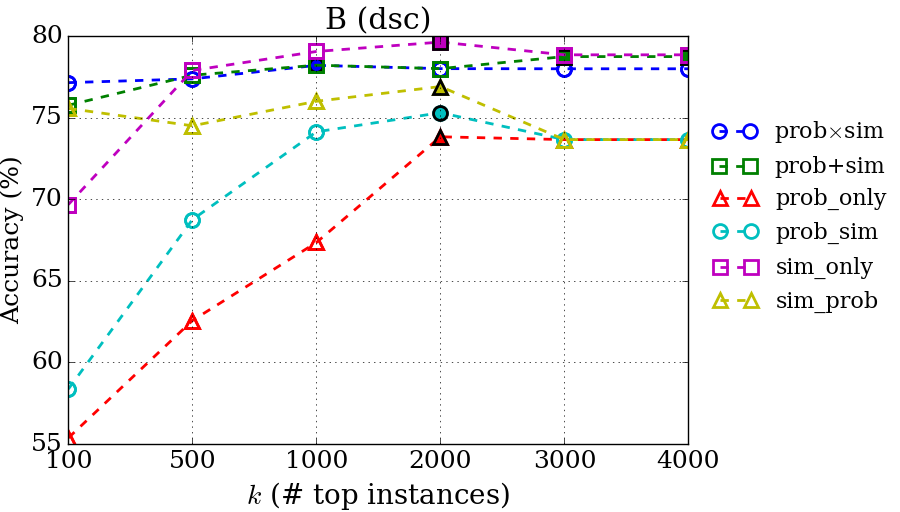}
	    \caption{prob sorted in descending order}
	    \label{fig:selection-methods-dsc}
	\end{subfigure}
	\caption{The number of selected pseudo-labelled instances $k$ on \textbf{Blitzer2007} is shown on the x-axis. prob denotes prediction confidence from the pseudo classifier trained on the source domains, sim denotes the similarity to the target domain, asc and dsc respectively denote sorted in ascending and descending order (only applied to prob related selection methods, sim is always sorted in dsc). prob\_only denotes using only prediction confidence, sim\_only denotes using only target similarity.
    prob\_sim indicates selecting by prob first and then sim (likewise for sim\_prob).
	prob$\times$sim denotes using the product of prob and sim, and prob+sim denotes using their sum. The marker for the best result of each method is filled.}
	\label{fig:pl-B}
\end{figure*}

\subsection{Pseudo-labelled Instances Selection}
When selecting the pseudo-labelled instances from the target domain for training a classifier for the target domain, we have two complementary strategies:
(a) select the most confident instances according to $f^{*}$ (denoted by \emph{prob}) or 
(b) select the most similar instances to the target domain's centroid (denoted by \emph{sim}).
To evaluate the effect of these two strategies and their combinations (i.e prob+sim and prob$\times$sim), in Figure~\ref{fig:pl-B}, we select target instances with each strategy and measure the accuracy on the target domain \textbf{B} for increasing numbers of instances $k$ in the descending (dsc) and ascending (asc) order of the selection scores.

From Figure~\ref{fig:selection-methods-dsc} we observe that selecting the highest confident instances does not produce the best UDA accuracies.
In fact, merely selecting instances based on confidence scores only (corresponds to prob\_only) reports the worst performance.
On the other hand, instances that are highly similar to the target domain's centroid are very effective for domain adaptation.
We observe that with only $k=1000$ instances, sim\_only reaches almost its optimal accuracy.
Using validation data, we estimated that $k=2000$ to be sufficient for all domains to reach the peak performance regardless of the selection strategy.
Therefore, we selected $2000$ pseudo-labelled instances for the attention step. 
In our experiments, we used sim\_only to select pseudo-labelled instances because it steadily improves the classification accuracy with  $k$ for all target domains, and is competitive against other methods.

\begin{table}[htb]
\centering
\resizebox{0.32\textwidth}{!}{
\begin{tabular}{@{}lllll@{}}
\toprule
T & uni-MS    & Self  & PL    & Att \\ \hline
B & 79.46 & 79.60  & 79.57 & \textbf{79.68}    \\
D & 82.32 & 82.49 & 82.71 & \textbf{82.96}    \\
E & 84.93 & 84.97 & \textbf{85.30}  & \textbf{85.30}    \\
K & 87.17 & 87.18 & 87.30  & \textbf{87.48}    \\ \hline
\end{tabular}
}
\caption{Classification accuracies (\%) across different steps of the proposed method, evaluated on \textbf{Chen2012}.}
\label{tbl:chen2012-steps}
\end{table}

\subsection{Effect of the Relatedness Map}\label{sec:experiments-map-att}

In Table~\ref{tbl:chen2012-steps}, we report the classification accuracy on the test instances in the target domain over the different steps: \textbf{uni-MS}~(no adapt baseline), \textbf{Self}~(self-training), \textbf{PL}~(pseudo-labelling) and \textbf{Att}~(attention).
We use the self-training method described in Algorithm~\ref{alg:MS-self-train}.
The results clearly demonstrate a consistent improvement over all the steps in the proposed method. 
For \textbf{Self} step, the proposed method improves the accuracy slightly without any information from the target domain. In the \textbf{PL} step, we report the results of a predictor trained on target pseudo-labelled instances. We report the evaluation results for the trained attention model in \textbf{Att}.

\begin{figure*}[t!]
	
	\begin{tabular}{l}
	
		\begin{minipage}[t]{1\textwidth}
			Example (1) \textit{Why anybody everest feet would want reading this? ... pure pleasure why 29028 feet account this?... It’s a pleasure to read.}
		\end{minipage}
	\end{tabular}
	\centering
	\begin{subfigure}[b]{0.32\textwidth}
		\includegraphics[width=\textwidth]{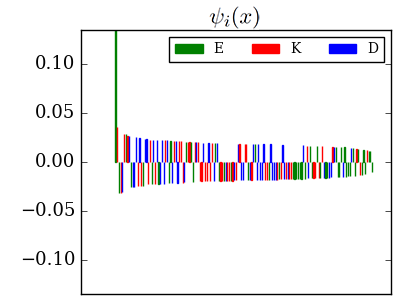}
		\caption{}\label{fig:ex1-psi}
		\vspace{-3mm}
	\end{subfigure}
	\begin{subfigure}[b]{0.32\textwidth}
		\includegraphics[width=\textwidth]{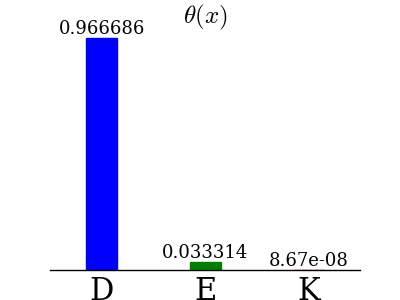}
		\caption{}\label{fig:ex1-theta}
		\vspace{-3mm}
	\end{subfigure}
	\begin{subfigure}[b]{0.32\textwidth}
		\includegraphics[width=\textwidth]{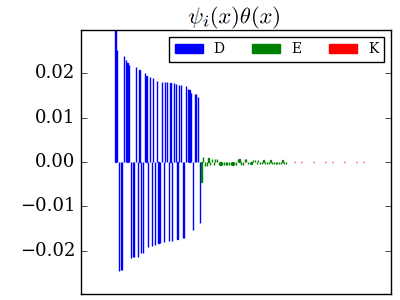}
		\caption{}\label{fig:ex1-all}
		\vspace{-3mm}
	\end{subfigure}
	\caption[A positively labelled a target test instance in B]{A positively labelled a target test instance in \textbf{B} (top) and resulted $\theta$, $\psi_i$ and the product of $\psi_i$ and $\theta$ (bottom). Here, the x-axis represents the instances and the y-axis represents the prediction scores. Instance specific values in (a) and (c) are shown as $>0$ for positive labelled instances and otherwise $<0$. Source instances from \textbf{D}, \textbf{E} and \textbf{K} are shown in blue, green and red respectively. The contributions from top-$150$ instances from three source domains are shown.}
	\label{fig:example-pos}
\end{figure*}

\begin{figure*}[t!]
	\begin{tabular}{l}
		\begin{minipage}[t]{1\textwidth}
			Example (2) \textit{Her relationship limited own pass her own analysis, there're issues mainly focus in turn for codependency. Disappointing, dysfunctional. Mother'll book her daughter's turn the pass, message turn the message issues analysis of very disappointing information.}
		\end{minipage}
	\end{tabular}
	\centering
	\begin{subfigure}[b]{0.32\textwidth}
		\includegraphics[width=\textwidth]{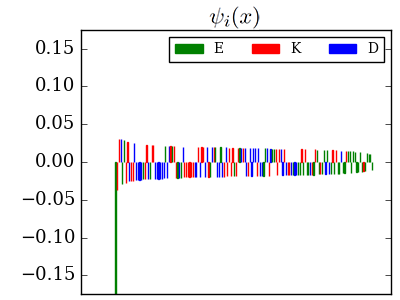}
		\caption{}\label{fig:ex2-psi}
		\vspace{-3mm}
	\end{subfigure}
	\begin{subfigure}[b]{0.32\textwidth}
		\includegraphics[width=\textwidth]{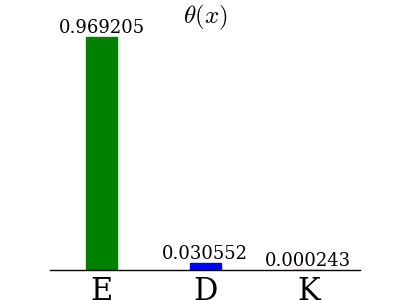}
		\caption{}\label{fig:ex2-theta}
		\vspace{-3mm}
	\end{subfigure}
	\begin{subfigure}[b]{0.32\textwidth}
		\includegraphics[width=\textwidth]{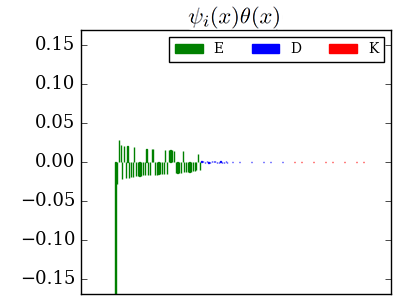}
		\caption{}\label{fig:ex2-all}
		\vspace{-3mm}
	\end{subfigure}
	\caption[A negatively labelled a target test instance in B]{A negatively labelled target test instance in \textbf{B}.}
	\vspace{-3mm}
	\label{fig:example-neg}
	\vspace{5mm}
	\resizebox{1\textwidth}{!}{
		\renewcommand{\arraystretch}{1.2}
		\begin{tabular}{cccl}
			\hline
			DM & L & Score   & Evidences (Reviews)                                                                                                                                 \\ \hline
			E      & -     & 0.16943 & \begin{minipage}[t]{0.75\textwidth}Serious problems.\end{minipage}                \\ \hline
			E      & -     & 0.02823 & \begin{minipage}[t]{0.75\textwidth}Sound great but lacking isolation in other areas.\end{minipage} \\ \hline
			E      & +     & 0.02801 & \begin{minipage}[t]{0.75\textwidth}Cases for the cats walking years, no around and knocking...walking on similar cases of cats.   \end{minipage}                     \\ \hline
			E      & +     & 0.02233 & \begin{minipage}[t]{0.75\textwidth}Cord supposed to no problems, this extension extension not worked as cord did...whatever expected just worked fine.\end{minipage}     \\ \hline
			E      & -     & 0.02209 & \begin{minipage}[t]{0.75\textwidth}Buy this like characters not used names...be aware of many commonly used characters before you accept file like drive.\end{minipage}     \\ \hline

		\end{tabular}
	}
	\captionof{table}[The top-$5$ evidences for Example (2)]{
		The top-$5$ evidences for Example (2) selected from the source domains.
		DM denotes the domain of the instance. L denotes the label for the instance. Score is $\psi_{i}(x)\theta(x)$.}
	\label{tbl:ex2-text}
\end{figure*}

In \textbf{Att} step, we use the relatedness map $\psi_{i}$ to express the similarity between a target instance and each of source domain instances, and the domain attention score $\theta$ to express the relation between a target instance and each of the source domain instances.
Two example test instances (one positive and one negative) from the target domain \textbf{B} are shown in Figure~\ref{fig:example-pos} and~\ref{fig:example-neg}. We observe that different source instances contribute to the predicted labels in different ways. 
As expected, in Figure~\ref{fig:ex1-psi} more positive source instances are selected using the relatedness map for a positive target instance, and Figure~\ref{fig:ex2-psi} more negative source instances are selected for a negative target instance. 
After training, we find that the proposed method identifies the level of importance of different source domains.
Example~(1) is closer to \textbf{D}, whereas Example~(2) is closer to \textbf{E} with a very high value of $\theta$. 
Figure~\ref{fig:ex1-all} and \ref{fig:ex2-all} show that the instance specific contribution to the target instance. We observe the proposed method also identifies the level of importance within the most relevant source domain. Table~\ref{tbl:ex2-text} shows the actual reviews as the top-$5$ evidences from the source domains in Example~(2). Negative labelled source training instance from \textbf{E}: \textit{``Serious problem.''} is the most important instance with the highest contribution of $\psi_{i}(x)\theta(x)$ to the decision. 


\section{Conclusions}
\label{sec:conclusion}
We propose a multi-source UDA method that combines self-training with an attention module. In contrast to prior works that select pseudo-labelled instances based on prediction confidence of a predictor learnt from source domains, our proposed method uses similarity to the target domain during adaptation. Our proposed method reports competitive performance against previously proposed multi-source UDA methods on two splits on a standard benchmark dataset.

\bibliographystyle{acl}
\bibliography{multi-source.bib}


\underline{\textbf{Multi-Source Attention for Unsupervised Domain Adaptation -- Supplementary Materials}}
 
\section*{Qualitative Analysis}
We show the actual reviews of the top-$k$ instances with high values according to $\psi_{i}(x)\theta(x)$. The target domain test instance and top-$5$ source domain instances are shown in Table~\ref{tbl:ex2-evd} for Example (1): a negatively labelled target test instance in \textbf{B}. 

\begin{table*}[htb]
\begin{tabular}{l}
\begin{minipage}[t]{1\textwidth}
Example (1) \textit{Why anybody everest feet would want reading this? ... pure pleasure why 29028 feet account this?... It’s a pleasure to read.}
\end{minipage}
\end{tabular}
\vspace{-5mm}
\end{table*}
\begin{table*}[htb]
\resizebox{1\textwidth}{!}{
\begin{tabular}{cccl}
\hline
DM & L & Score   & Evidence (Reviews)                                                                                                                                 \\ \hline
D      & +     & 0.02981 & \begin{minipage}[t]{0.75\textwidth} Children seeing what happened... best figures for warning this 911 happened real destruction...authority documentary.\end{minipage}                \\ \hline
D      & +     & 0.02531 & \begin{minipage}[t]{0.75\textwidth}Blind strength for negligence a lump justice and against himself...no justice shall be a system against great and greater odds words.\end{minipage} \\ \hline
D      & -     & 0.02459 & \begin{minipage}[t]{0.75\textwidth}Pathetic feel tawdry pathetic moments, wants to only to later...but clear later or greatest a week fact once.   \end{minipage}                     \\ \hline
D      & +     & 0.02399 & \begin{minipage}[t]{0.75\textwidth}Ties of hurt and gripping it poverty who cannot decides to see this...this film defeats its path...takes destroy of life.\end{minipage}     \\ \hline
D      & +     & 0.02301 & \begin{minipage}[t]{0.75\textwidth}He believes the worst day is our history, terrorist attack reviewer...should be furthest day from attack, never be an American.\end{minipage}     \\ \hline
%
\end{tabular}
}
\caption{The top-$5$ evidences for Example (1) selected from the source domains.
DM denotes the domain of the instance. L denotes the label for the instance. Score is $\psi_{i}(x)\theta(x)$.}
\label{tbl:ex2-evd}
\end{table*}

\section*{Pseudo-labelled Instances Selection}\label{sup:pseudo-label}
\begin{figure}[ht!]
\centering
	\includegraphics[width=0.45\textwidth]{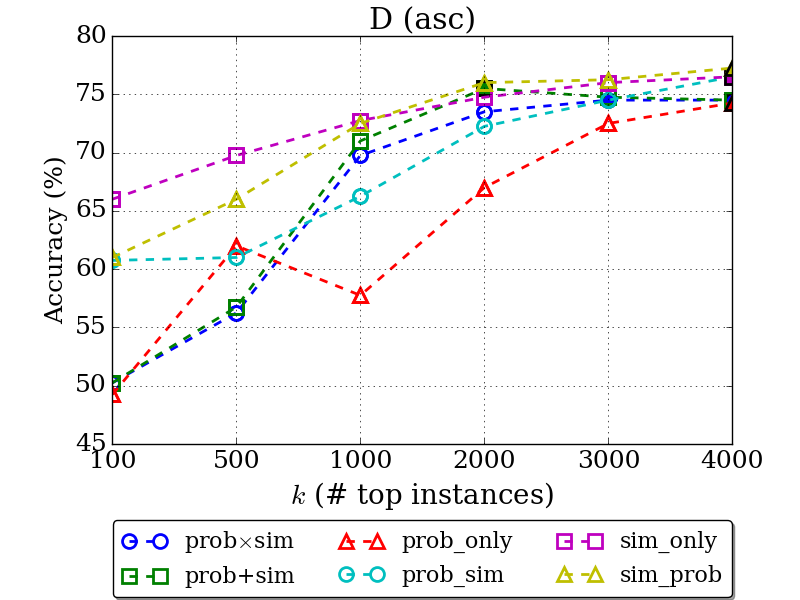}
	\includegraphics[width=0.45\textwidth]{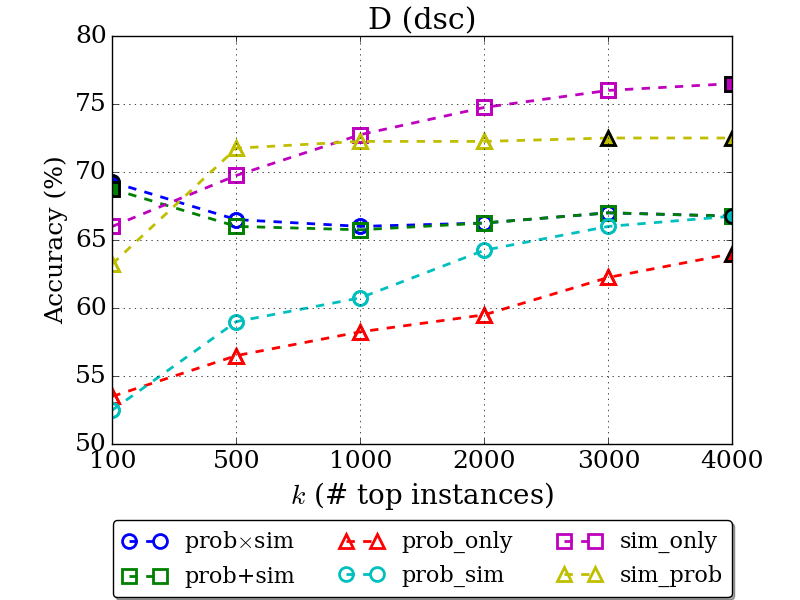}
	\includegraphics[width=0.45\textwidth]{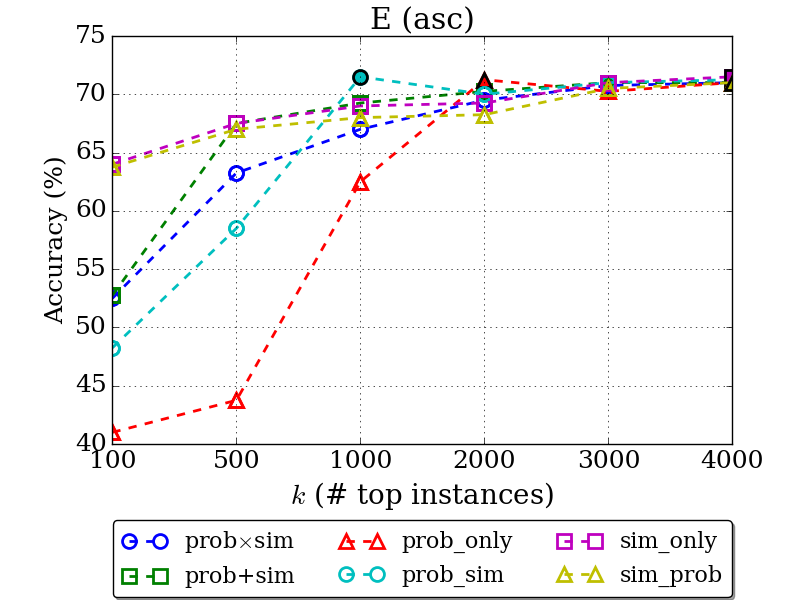}
	\includegraphics[width=0.45\textwidth]{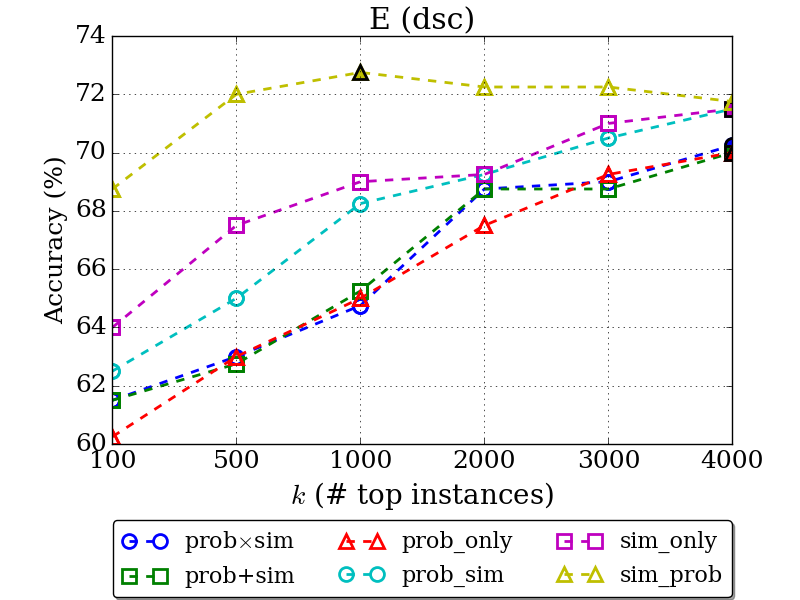}
	\includegraphics[width=0.45\textwidth]{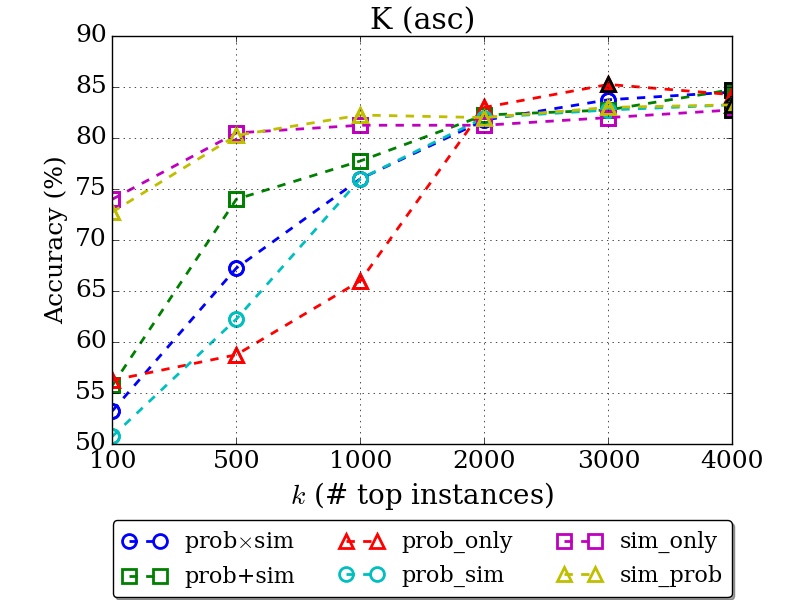}
	\includegraphics[width=0.45\textwidth]{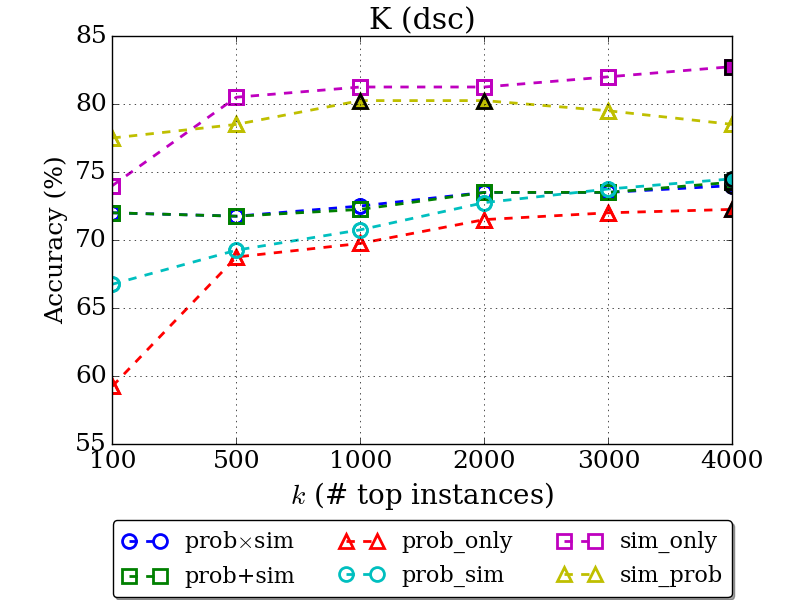}
	\caption{The number of selected pseudo-labelled instances $k$ on \textbf{Blitzer2007} is shown on the x-axis. prob denotes prediction confidence from the pseudo classifier trained on the source domains, sim denotes the similarity to the target domain, asc and dsc respectively denote sorted in ascending and descending order (only applied to prob related selection methods, sim is always sorted in dsc). prob\_only denotes using only prediction confidence, sim\_only denotes using only target similarity.
    prob\_sim indicates selecting by prob first and then sim (likewise for sim\_prob).
	prob$\times$sim denotes using the product of prob and sim, and prob+sim denotes using their sum. The marker for the best result of each method is filled.}\label{fig:pl-all-sup}
\end{figure}

In Figure~\ref{fig:pl-all-sup}, we report the results for the \textbf{PL} step when different selection criteria are used on all target domains in \textbf{Blitzer2007}.

\end{document}